\documentclass[conference]{IEEEtran}
\usepackage{amsmath,amsfonts}
\usepackage{algorithmic}
\usepackage{array}
\usepackage[caption=false,font=normalsize,labelfont=sf,textfont=sf]{subfig}
\usepackage{textcomp}
\usepackage{stfloats}
\usepackage{url}
\usepackage{verbatim}
\usepackage{graphicx}
\usepackage{tabularx}
\usepackage{enumitem}
\usepackage{color}
\usepackage[table]{xcolor}
\usepackage{booktabs}
\usepackage{amssymb}
\usepackage{multicol,multirow}
\usepackage{bm}
\usepackage{tikz,tkz-euclide}
\usepackage{bbm}
\usepackage{threeparttable}
\usepackage[hidelinks]{hyperref}
\usetikzlibrary{fit,arrows,shapes,quotes,angles,calc,decorations.pathreplacing}

\usepackage[colorinlistoftodos]{todonotes}

\begin{document}

\title{Stack more levels: how to get general and human-like Mario playing}

\author{
\IEEEauthorblockN{Shinichi Arimura, Yuchen Li, and Julian Togelius}
\IEEEauthorblockA{New York University\\
Brooklyn, New York, USA\\
sa9090@nyu.edu, yl6394@nyu.edu, julian@togelius.com}
}

\IEEEoverridecommandlockouts

\IEEEpubid{%
\makebox[\columnwidth]{979-8-3315-9476-3/26/\$31.00~\copyright~2026 IEEE\hfill}%
\hspace{\columnsep}\makebox[\columnwidth]{}%
}

\maketitle

\IEEEpubidadjcol

\begin{abstract}
Super Mario Bros. is widely used in game AI and procedural content 
generation (PCG), yet it remains unclear whether a model-free agent can 
reliably win unseen levels. To address this question, we train Proximal 
Policy Optimization (PPO) agents on increasingly large sets of 
PCG-generated Mario levels and evaluate their generalization on 1000 
held-out test levels. We then shape competent PPO policies into three 
gameplay archetypes---runner, killer, and collector. Finally, we 
fine-tune these PPO policies using Diffusion-Reward Adversarial 
Imitation Learning (DRAIL) with human gameplay demonstrations. We find that increasing the number of distinct PCG-generated training levels generally improves unseen-level performance, and that fine-tuning PPO policies with DRAIL increases action-level human-likeness while preserving much of the persona induced by PPO pretraining.
\end{abstract}

\begin{IEEEkeywords}
game AI, reinforcement learning, Super Mario Bros., human-like agents, playstyle
\end{IEEEkeywords}

\section{Introduction}\label{sec:intro}
Perhaps surprisingly, it remains unclear whether any agent can reliably play Super Mario Bros in a real-time model-free setting and win unseen levels. There are strong planning agents (such as the ubiquitous A* agent by Baumgarten~\cite{karakovskiy2012mario}), but those require access to a fast forward model, something that most games do not give you. There are also agents that can play individual levels well, i.e., agents that overfit~\cite{togelius2009super}.

Super Mario Bros (SMB), in the form of the Mario AI framework~\cite{karakovskiy2012mario}, is a widely used benchmark, and a competent player agent underpins much PCG and player-modeling research~\cite{shaker20112010,summerville2018pcgml,pedersen2010modeling}. As an archetypical side-scrolling platformer, SMB also exposes a question that remains open: what it takes to play even a ``simple'' game well across unseen levels~\cite{volz2018evolving,shu2021experience}. This concern is not specific to SMB; much reinforcement learning research, including work on the Atari ALE framework~\cite{bellemare2013arcade}, evaluates agents that overfit to a small set of levels~\cite{machado2018revisiting,cobbe2019quantifying,cobbe2020leveraging}.

Here, we start by asking the simple question if we can train a model-free reinforcement learning agent to play Mario well on unseen levels, and if so, how many levels it must be trained on. After establishing that this is possible (mostly by adding more levels to the training set), we investigate if we can get them to play in different human-like ways. Finally, we investigate whether we can make these policies more human-like by fine-tuning them with imitation learning~\cite{hussein2017imitation}. In this work, we use "human-like" in a narrow, action-level sense: a policy is more human-like if, on held-out human states, its action choices agree more often with the recorded human action and its marginal action distribution is closer to that of human players. We do not claim human-likeness at the level of strategy, intent, or subjective perception.
Our contributions are: (i) training-pool size strongly affects unseen-level performance for model-free Mario agents, and (ii) DRAIL fine-tuning increases action-level human-likeness while preserving the persona learned during PPO pretraining.

\section{Related Work}

\textbf{Mario agents and model-free generalization.}
The Mario AI benchmark has a long history of strong search- and planning-based agents~\cite{togelius20102009,togelius2013mario}. However, such agents often rely on a fast forward model, whereas our agents are model-free policies that act directly from observations. We use Proximal Policy Optimization (PPO), a widely used policy-gradient algorithm~\cite{schulman2017proximal}, placing our work closer to RL generalization studies in benchmarks such as ALE, where overfitting to narrow environment distributions is a central concern~\cite{bellemare2013arcade}.


\textbf{Human-like and playstyle agents.}
Recent work has explored controllable or human-like game agents through interactive reinforcement learning, behavior cloning, adversarial imitation, and macro-action abstraction~\cite{arzate2020mariomix,de2022automated,ingram2023creating,ahlberg2023generating,guo2025learning}. Behavior cloning predicts human actions from demonstrations, but can suffer from rollout distribution shift. Adversarial imitation methods such as Generative Adversarial Imitation Learning (GAIL) instead match expert and agent state-action distributions through interaction~\cite{ho2016generative}.  Diffusion-Reward Adversarial Imitation
Learning (DRAIL) extends this idea with a diffusion-based discriminator that provides smoother imitation rewards, making it suitable as a post-training shaping signal for competent PPO personas~\cite{lai2024diffusion}.

\section{Method}

Our method has three stages: PPO training for level generalization, segment-reward shaping for playstyles, and DRAIL fine-tuning with human demonstrations.

\subsection{Training Levels and Evaluation Split}

All levels are generated using Experience-Driven PCG via Reinforcement Learning~\cite{shu2021experience} and split into disjoint training and test sets. We filter the test set to levels with more than 10 coins and more than 10 enemies, yielding 1000 held-out levels with enough opportunities for collection and enemy interaction.

To study level quantity, we create four fixed training pools with $N\in\{100,1000,10000,50000\}$ distinct PCG-generated levels. PPO episodes are sampled using inverse-visit weighting, so less-visited levels are sampled more often. All agents share the same architecture and PPO configuration; only pool size and playstyle reward differ.

\subsection{Segment-Based Playstyle Rewards}

We train three playstyle archetypes: runner, killer, and collector. The runner is encouraged to complete levels directly, the killer to engage enemies, and the collector to gather coins and power-ups. Instead of dense horizontal progress, we use 16-tile segment checkpoints, preserving completion while allowing local detours for collection or enemy engagement.
All agents share the same base competence reward:
\begin{equation}
r_t^{\mathrm{base}}
=
w_{\mathrm{seg}} \Delta S_t
+
w_{\mathrm{win}}\mathbbm{1}[\mathrm{WIN}_t].
\end{equation}
Here, $\Delta S_t$ is the number of newly reached 16-tile segment checkpoints at time $t$, and $\mathbbm{1}[\mathrm{WIN}_t]$ indicates whether the episode ends in a win. Each segment gives reward only the first time it is reached.

The final reward is the base and a style-specific bonus:
\begin{align}
r_t^{\mathrm{runner}}
&=
r_t^{\mathrm{base}}, \\
r_t^{\mathrm{collector}}
&=
r_t^{\mathrm{base}}
+
w_c\Delta C_t
+
w_m\Delta M_t
+
w_f\Delta F_t, \\
r_t^{\mathrm{killer}}
&=
r_t^{\mathrm{base}}
+
w_k \Delta K_t^* .
\end{align}
Here, $\Delta C_t$, $\Delta M_t$, and $\Delta F_t$ denote newly collected coins, mushrooms, and fire flowers, and $\Delta K_t^*$ denotes newly credited enemy kills, capped per segment to prevent farming. We set $w_{\mathrm{seg}}=1.0$, $w_{\mathrm{win}}=5.0$, and $w_c=w_m=w_f=w_k=2.5$.

Thus, all agents share the same completion objective, but differ in their secondary incentives. The runner uses no extra time reward because, without secondary objectives, detours do not increase return.

\subsection{DRAIL Fine-Tuning}

Reward-shaped PPO agents can be competent and stylistically distinct, but still non-human-like. We fine-tune each policy with DRAIL~\cite{lai2024diffusion}, whose discriminator $D_\phi$ predicts whether $(o_t,a_t)$ comes from human demonstrations or the current policy. The imitation reward is
\begin{equation}
    r_t^{\mathrm{DRAIL}} =
    \log D_\phi(o_t,a_t) - \log(1-D_\phi(o_t,a_t)).
\end{equation}
Let $r_t^{\mathrm{env}}$ be the original PPO persona reward and let 
$\lambda_{\mathrm{DRAIL}}$ weight the imitation reward. During PPO$\rightarrow$DRAIL 
fine-tuning, we use
\begin{equation}
    r_t^{\mathrm{FT}} =
    r_t^{\mathrm{env}} + \lambda_{\mathrm{DRAIL}} r_t^{\mathrm{DRAIL}}.
\end{equation}
We set $\lambda_{\mathrm{DRAIL}}=1.2$, $2.9$, and $2.5$ for runner, killer, and collector, selected independently via an Optuna sweep optimizing Action Agreement Ratio (AAR).

\subsection{Human Demonstration Split}

The human demonstration pool is taken from the human gameplay study of~\cite{wang2025fun} and contains 220 action-state trajectories collected under a five-life protocol. For DRAIL training, we select 50 style-filtered trajectories per archetype: short-run trajectories for runner, high-kill trajectories for killer, and high-coin trajectories for collector. These sets can overlap, since a trajectory may satisfy more than one criterion; the remaining 95 unique trajectories are held out for evaluation.
We evaluate Direct DRAIL and PPO$\rightarrow$DRAIL rollouts with five lives to match the demonstration collection protocol.

\section{Experiments and Results}

We evaluate three questions: (i) does increasing the number of training levels improve unseen-level generalization, (ii) do segment rewards produce distinct playstyles, and (iii) does PPO$\rightarrow$DRAIL fine-tuning improve human-likeness?

\subsection{Metrics and Statistical Tests}

Competence is measured by win rate and completion. For PPO level-scaling, we use a one-life, best-of-10 protocol: each level is counted as won if any of 10 stochastic rollouts reaches the goal, and completion is the maximum progress across rollouts. This measures per-level capability under repeated attempts, not single-rollout success. DRAIL rollouts use five lives to match the human demonstrations.

Playstyle is measured by a target metric for each archetype: episode length for runner, kill ratio for killer, and coin ratio for collector. For runner, fewer steps indicate shorter episodes, interpreted together with win rate and completion to avoid rewarding fast failure. Kill ratio is killed enemies divided by the number of enemies initially present in the level, and coin ratio is collected coins divided by total coins in the level. Kill ratio can exceed 1 because enemies may respawn.

Human-likeness is measured by Action Agreement Ratio (AAR), following Holmg{\aa}rd et al.'s decision-level comparison of agent and human play styles~\cite{holmgard2014personas}. AAR is the fraction of held-out human states where the agent's most likely action matches the human action; higher AAR indicates stronger action-level agreement. Avg. AAR and Avg. Action JS are the simple averages over the runner, killer, and collector policies.

\begin{table}[t]
\centering
\caption{
One-life PPO evaluation on 1000 held-out test levels.
}
\label{tab:ppo_results}
\small
\setlength{\tabcolsep}{4pt}
\begin{threeparttable}
\begin{tabular}{llcccc}
\toprule
Agent & Levels & Win & Compl. & Kill & Coin \\
\midrule
\multirow{4}{*}{Runner}
& 100 & 0.091 & 0.415 & 0.092 & 0.125 \\
& 1k  & 0.462 & 0.709 & 0.135 & 0.238 \\
& 10k & 0.630 & 0.806 & 0.142 & 0.248 \\
& 50k & \textbf{0.648} & \textbf{0.817} & 0.151 & 0.253 \\
\midrule
\multirow{4}{*}{Killer}
& 100 & 0.222 & 0.561 & 0.384 & 0.158 \\
& 1k  & 0.540 & 0.776 & 0.638 & 0.221 \\
& 10k & \textbf{0.823} & \textbf{0.926} & \textbf{0.846} & 0.307 \\
& 50k & 0.680 & 0.832 & 0.747 & 0.247 \\
\midrule
\multirow{4}{*}{Collector}
& 100 & 0.202 & 0.549 & 0.174 & 0.382 \\
& 1k  & 0.535 & 0.786 & 0.267 & 0.653 \\
& 10k & 0.751 & 0.888 & 0.326 & 0.804 \\
& 50k & \textbf{0.855} & \textbf{0.929} & 0.321 & \textbf{0.845} \\
\bottomrule
\end{tabular}
\end{threeparttable}
\end{table}
For win rate, we fit a logistic mixed-effects model with training-pool size, playstyle, and their interaction as fixed effects, and random intercepts for level and seed. Completion and playstyle contrasts use paired bootstrap CIs over levels, with Holm-corrected Wilcoxon signed-rank tests for playstyle $p$ values. AAR uses paired bootstrap CIs over the 95 held-out human trajectories.

\subsection{PPO Level Generalization and Playstyle}

Table~\ref{tab:ppo_results} reports the one-life PPO level-scaling evaluation. Increasing the number of training levels generally improves unseen-level win rate and completion. Runner improves from 9.1\% to 64.8\% and collector from 20.2\% to 85.5\%, the best final win rate, while killer peaks at 82.3\% (10k) before dropping to 68.0\% (50k).

\begin{table}[t]
\centering
\caption{One-life generalization statistics on held-out levels.}
\label{tab:generalization_stats}
\small
\setlength{\tabcolsep}{1.5pt}
\renewcommand{\arraystretch}{0.95}
\begin{threeparttable}
\begin{tabular}{@{}lccc@{}}
\toprule
Agent 
& OR$_{10\times}$ Win
& 10k$\rightarrow$50k Win
& 10k$\rightarrow$50k Compl. \\
\midrule
Runner 
& 3.70 [3.65, 3.74]
& OR 1.12, $p=.099$
& +0.011 [-0.014, 0.034] \\
Killer 
& 2.83 [2.76, 2.90]
& OR 0.37, $p<.001$
& -0.094 [-0.112, -0.075] \\
Collector 
& 4.34 [4.23, 4.46]
& OR 2.30, $p<.001$
& +0.041 [0.026, 0.057] \\
\bottomrule
\end{tabular}
\vspace{0.25em}
\begin{tablenotes}
\footnotesize
\item OR$_{10\times}$ is per tenfold level increase; 10k$\rightarrow$50k is the final adjacent contrast.
\end{tablenotes}
\end{threeparttable}
\end{table}

The logistic mixed-effects model confirms a significant positive effect of training-level count on unseen-level win probability. Each tenfold increase multiplies the odds of winning by 3.70 for runner, 2.83 for killer, and 4.34 for collector. To avoid forcing a log-linear trend, we also fit a categorical-pool win model and test adjacent training-pool contrasts. The 10k to 50k contrast shows a playstyle-dependent pattern: runner  plateaus, killer decreases, and collector continues improving. Completion shows the same qualitative pattern under paired bootstrap. We therefore interpret the result as positive scaling overall with no single universal saturation point.
One possible explanation for the killer drop is that all pool sizes were trained for the same total number of timesteps; with 50k levels, each individual level is visited less often, and the riskier enemy-engagement behavior may require more samples to stabilize than running or collecting.

Table~\ref{tab:style_stats} confirms that the final 50k agents express distinct playstyles. Runner has significantly shorter episodes than killer and collector, killer has significantly higher kill ratio than both runner and collector, and collector has significantly higher coin ratio than both runner and killer. All six planned Wilcoxon tests remain significant after Holm correction.

\begin{table}[t]
\centering
\caption{
Playstyle differentiation at the final 50k PPO checkpoint.
}
\label{tab:style_stats}
\small
\begin{threeparttable}
\begin{tabularx}{\columnwidth}{Xccc}
\toprule
Comparison & Mean Diff. & 95\% CI & Holm $p$ \\
\midrule
K $-$ R steps & 116.737 & [96.796, 137.175] & $<10^{-49}$ \\
C $-$ R steps & 187.274 & [165.583, 210.795] & $<10^{-90}$ \\
K $-$ R kill & 0.598 & [0.578, 0.619] & $<10^{-150}$ \\
K $-$ C kill & 0.425 & [0.401, 0.448] & $<10^{-123}$ \\
C $-$ R coin & 0.593 & [0.575, 0.612] & $<10^{-158}$ \\
C $-$ K coin & 0.598 & [0.580, 0.616] & $<10^{-159}$ \\
\bottomrule
\end{tabularx}

\vspace{0.5em}
\begin{tablenotes}
\footnotesize
\item R = Runner, K = Killer, C = Collector.
\end{tablenotes}
\end{threeparttable}
\end{table}

These results show that segment-based rewards induce distinct playstyles while preserving the shared goal of level completion. Runner shortens episodes, killer increases enemy engagement, and collector increases resource collection.

\begin{table*}[t]
\centering
\caption{
Human-likeness on 95 held-out human trajectories.
}
\label{tab:aar_js}
\small
\begin{threeparttable}
\begin{tabular}{lccccc}
\toprule
Method 
& Runner AAR 
& Killer AAR 
& Collector AAR 
& Avg. AAR 
& Avg. Action JS \\
\midrule
PPO 
& 0.016 
& 0.007 
& 0.021 
& 0.015 
& 0.643 \\
Direct DRAIL 
& 0.355 {\scriptsize(+0.339 [0.322, 0.356])}
& 0.416 {\scriptsize(+0.409 [0.394, 0.424])}
& 0.420 {\scriptsize(+0.399 [0.379, 0.417])}
& 0.397
& 0.103 \\
PPO$\rightarrow$DRAIL 
& 0.387 {\scriptsize(+0.371 [0.353, 0.388])}
& 0.192 {\scriptsize(+0.185 [0.172, 0.198])}
& 0.393 {\scriptsize(+0.372 [0.354, 0.388])}
& 0.324
& 0.218 \\
\bottomrule
\end{tabular}
\end{threeparttable}
\end{table*}

\subsection{DRAIL and Human-Likeness}

Table~\ref{tab:aar_js} reports human-likeness on the 95 held-out human trajectories. AAR measures exact agreement with human actions, while Action JS measures Jensen--Shannon divergence between the marginal action distributions of the agent and human data. Thus, higher is better for AAR, while lower is better for Action JS. Avg. AAR and Avg. Action JS are computed as the simple averages of the corresponding runner, killer, and collector scores.


The original PPO agents have near-zero AAR, showing that strong reward-shaped policies are not action-level human-like. Direct DRAIL achieves the strongest pure imitation score, with Avg. AAR 0.397 and Avg. Action JS 0.103. PPO$\rightarrow$DRAIL also substantially improves over PPO, reaching Avg. AAR 0.324 and reducing Avg. Action JS from 0.643 to 0.218. The paired trajectory-level bootstrap confirms that PPO$\rightarrow$DRAIL significantly improves AAR over PPO for all three roles. The improvement is strongest for runner and collector, while the killer improvement is smaller.

\begin{table*}[t]
\centering
\caption{
Five-life rollout evaluation for persona preservation.
}
\label{tab:drail_persona}
\small
\setlength{\tabcolsep}{4pt}
\begin{threeparttable}
\begin{tabular}{lccc|ccc|ccc}
\toprule
\multirow{2}{*}{Method}
& \multicolumn{3}{c|}{Runner}
& \multicolumn{3}{c|}{Killer}
& \multicolumn{3}{c}{Collector} \\
\cmidrule(lr){2-4}
\cmidrule(lr){5-7}
\cmidrule(lr){8-10}
& Win & Compl. & Steps
& Win & Compl. & Kill
& Win & Compl. & Coin \\
\midrule
PPO
& 0.995 & 0.998 & 435.0
& 0.985 & 0.993 & 1.023
& 0.990 & 0.996 & 0.927 \\

Direct DRAIL
& 0.922 & 0.968 & 1042.1
& 0.927 & 0.973 & 0.853
& 0.904 & 0.967 & 0.565 \\

PPO$\rightarrow$DRAIL
& 0.971 & 0.984 & 915.2
& 0.968 & 0.989 & 1.109
& 0.990 & 0.994 & 0.915 \\
\bottomrule
\end{tabular}
\end{threeparttable}
\end{table*}

Table~\ref{tab:drail_persona} shows why imitation metrics are insufficient. Direct DRAIL achieves the best AAR and Action JS, but struggles to produce distinct rollout personas because the human demonstrations are not clearly separated by style. PPO$\rightarrow$DRAIL starts from a reward-shaped PPO policy and better preserves target behaviors: runner remains faster, killer kill ratio improves from 0.853 to 1.109, and collector coin ratio improves from 0.565 to 0.915, providing a practical compromise between persona preservation and human-likeness.



\section{Conclusion and Future Work}

Our results suggest three lessons. First, training-level quantity is important for model-free Mario competence: more distinct PCG-generated levels improve generalization to 1000 held-out levels, although the effect is playstyle-dependent. Collector scales most strongly, while killer peaks at 10k levels, suggesting that enemy-focused behavior is more sensitive to the training-level distribution and total training budget.

Second, segment-based rewards induce distinct playstyles while preserving the shared completion objective. Rewarding progress only at 16-tile checkpoints avoids constantly pushing the policy rightward, leaving room for collector and killer behavior within each segment. The final PPO agents remain strongly separated across all six planned playstyle contrasts.

Third, human demonstrations are useful but noisy. Direct DRAIL achieves the strongest  imitation metrics, but imitation alone does not recover task competence or persona behavior. PPO$\rightarrow$DRAIL starts from a competent reward-shaped policy and fine-tunes toward human-like actions while retaining more persona, suggesting a practical pipeline: train for competence, shape playstyle, then fine-tune for human-likeness.

The main limitations are that we vary training-level quantity rather than explicitly measuring level diversity, and that human-likeness evaluation uses a small demonstration pool and exact-action AAR, which may under-credit different but plausible human actions. Future work will use explicit level-diversity measures, and larger human datasets.

\bibliographystyle{IEEEtran}
\bibliography{biblography}

\end{document}